\documentclass[times,twocolumn,final]{elsarticle}

\usepackage{framed,multirow}

\usepackage{amssymb}
\usepackage{latexsym}
\usepackage{listings}

\usepackage{url}
\usepackage{xcolor}
\usepackage[hidelinks]{hyperref}
\definecolor{newcolor}{rgb}{.8,.349,.1}

\usepackage[ruled,vlined,algo2e]{algorithm2e}
\SetKwComment{Comment}{$\triangleright$\ }{}

\usepackage{mathtools}
\usepackage{amsmath}
\usepackage{hyperref}
\usepackage{booktabs}
\usepackage{subcaption}

\newcommand{\rev}[1]{\textcolor{black}{#1}}
\begin{document}

\begin{frontmatter}

\title{FairREAD: Re-fusing Demographic Attributes after Disentanglement for Fair Medical Image Classification}

\author[1]{Yicheng Gao}
\ead{yicheng.gao1@northwestern.edu}
\author[1]{Jinkui Hao}
\author[1]{Bo Zhou}
\ead{bo.zhou@northwestern.edu}

\address[1]{Department of Radiology, Northwestern University, Chicago, IL, USA}

\begin{abstract}

Recent advancements in deep learning have shown transformative potential in medical imaging, yet concerns about fairness persist due to performance disparities across demographic subgroups. Existing methods aim to address these biases by mitigating sensitive attributes in image data; however, these attributes often carry clinically relevant information, and their removal can compromise model performance—a highly undesirable outcome. To address this challenge, we propose Fair Re-fusion After Disentanglement (FairREAD), a novel, simple, and efficient framework that mitigates unfairness by re-integrating sensitive demographic attributes into fair image representations. FairREAD employs orthogonality constraints and adversarial training to disentangle demographic information while using a controlled re-fusion mechanism to preserve clinically relevant details. Additionally, subgroup-specific threshold adjustments ensure equitable performance across demographic groups. Comprehensive evaluations on a large-scale clinical X-ray dataset demonstrate that FairREAD significantly reduces unfairness metrics while maintaining diagnostic accuracy, establishing a new benchmark for fairness and performance in medical image classification. \rev{Our code is available at: \url{https://github.com/Advanced-AI-in-Medicine-and-Physics-Lab/FairREAD/}}

\end{abstract}

\end{frontmatter}

\newcommand{\RR}{\mathbb{R}}
\newcommand{\dz}{{d_z}}
\newcommand{\da}{{d_a}}
\newcommand{\za}{{z_A}}
\newcommand{\zt}{{z_T}}
\newcommand{\Za}{{Z_A}}
\newcommand{\Zt}{{Z_T}}
\newcommand{\Zarow}{\Hat{Z_A^i}}
\newcommand{\Ztrow}{\Hat{Z_T^i}}
\newcommand{\Zacol}{\Tilde{Z_A^i}}
\newcommand{\Ztcol}{\Tilde{Z_T^i}}
\newcommand{\fa}{{f_A}}
\newcommand{\ft}{{f_T}}
\newcommand{\fadv}{f_{adv}}
\newcommand{\phia}{\phi_a}
\newcommand{\phit}{\phi_t}
\newcommand{\heada}{h_a}
\newcommand{\headt}{h_t}
\renewcommand{\L}{\mathcal{L}}

\section{Introduction}

With the rapid development of deep learning-based computer vision and medical imaging techniques, artificial intelligence (AI) models are receiving increasingly widespread attention for their enormous potential in diagnosis and assisting medical decisions. As of August 2024, the U.S. Food and Drug Administration (FDA) has approved 950 AI/ML-enabled medical devices, with the list expanding month by month \citep{FDAlist}. However, concerns about the fairness of these models have surfaced, as studies~\citep{Gichoya2022, Glocker2023, ShortcutEncode} reveal that deep learning models can inherently encode sensitive information such as race, gender, and age, which can lead to significant performance disparities across various demographic subgroups. This phenomenon, referred to as ``unfairness," poses challenges to equitable healthcare delivery.

In the context of deep learning, unfairness is formally defined as ``the phenomenon where the effectiveness of deep learning models notably favors or opposes one subgroup over another" \citep{xu2024addressing}. Unfairness has been widely observed in medical imaging models. For instance, \rev{\citet{unfairness-cause-dataset-2},} \citet{Chexclusion} and \citet{chestxray-fairness-zhang} examined chest X-ray classifiers and observed significant performance gaps across sex, age, and ethnicity subgroups, measured by metrics such as True Positive Rate (TPR). Similar disparities have been found in tasks such as brain MRI reconstruction \citep{mri-fairness}, cardiac MRI segmentation \citep{cardiac-mri-seg-fairness}, and thoracic CT classification \citep{ct-fairness}. \rev{Other works such as \citet{fairmedfm} propose benchmarks that facilitate detection and mitigation of bias in medical imaging models.}

The underlying causes of unfairness in medical imaging models remain an open question. Some studies suggest that unequal representation of subgroups in training datasets might be the cause \citep{unfairness-cause-dataset-1, unfairness-cause-dataset-2}. However, \citet{Chexclusion} found no statistical significance in the correlation between subgroup proportion and model performance disparity. \citet{chestxray-fairness-zhang} showed that one possible cause of unfairness is the disparity in label accuracy for different subgroups. Additionally, a line of research proposes that unfairness might be caused by spurious correlations, or ``shortcuts," learned by models during the training process \citep{banerjee2023shortcuts}.

To mitigate the unfairness problem in deep learning, various methods have been proposed, which can be broadly classified into pre-processing, in-processing, and post-processing techniques \citep{LASurvey, NBESurvey}. Here we provide a brief review of the existing approaches.

\textbf{Pre-processing} methods involve modifying the training data before model training. \textbf{Dataset re-balancing} methods such as oversampling underrepresented subgroups \citep{resample} or generating synthetic data \citep{paul2022tara} aim to balance the dataset distribution. While effective for addressing class imbalance, these techniques risk overfitting to minority class samples or introducing synthetic biases.
\textbf{Perturbation} methods aim to directly alter the distribution of one or more variables in the training data. This approach is akin to "repairing" certain aspects of the data to enhance fairness. Applications of perturbation-based data repair~\citep{johndrow2019algorithm,jiang2020identifying} have demonstrated that accuracy is not significantly impacted.
\textbf{Causal} methods \citep{salimi2019interventional} may help to identify dependency information and modify training data related to sensitive attributes. These methods provide robust insights but rely on domain knowledge and computationally expensive models.

\textbf{Post-processing} methods adjust the model's outputs to reduce biases~\citep{lepri2018fair}.
\textbf{Thresholding} approaches~\citep{hardt2016equality} aim to reduce performance disparities across subgroups by separately fitting classification thresholds for each subgroup. While sometimes effective, these methods suffer from bad generalizability, as they could lead to extreme classification thresholds that significantly \rev{impact} classification performance.
\textbf{Calibration} is the process of ensuring that predicted probabilities are well-calibrated across subgroups helps mitigate bias in decision-making~\citep{noriega2019active}, which is particularly useful when subgroup distributions differ significantly. \textbf{Pruning} methods measure the importance of each neuron in encoding sensitive attributes, and selectively remove neurons that contribute to unfairness. For example, \cite{wu2022fairpruneachievingfairnesspruning} measures the difference in saliency of each neuron when classifying patients with dark and light skins, and remove the neurons that demonstrate the most significant disparities. \cite{jung2024unifieddebiasingapproachvisionlanguage} trains a random forest model to predict sensitive attributes from each image feature, and remove the features with the highest importance values. Post-processing approaches are applicable to models after the training process is complete, without the need to rerun the training process. However, the artificial removal of model parameters can damage certain features and lead to significant performance loss.

Unlike pre- and post-processing methods, \textbf{in-processing} techniques directly incorporate fairness consideration during the model design phase~\citep{wan2023processing}, ensuring inherently fair models and addressing fairness issues at their core. \textbf{Adversarial learning} involves training a classification model alongside an adversarial model tasked with predicting demographic attributes. By minimizing the adversarial model's accuracy using a reverse gradient, the diagnostic model learns to suppress sensitive information~\citep{zhao2020training,bevan2022detecting,li2023enhancing}. However, the adversarial learning process can be unstable and difficult to converge. \textbf{Disentanglement} methods aim to separate information related to demographic attributes from the target representation used for disease prediction. This can be achieved by maximizing entropy, minimizing conditional mutual information, or encouraging orthogonality between sensitive attribute prediction and target classification \citep{FairOrth, Orth,vento2022penalty,more2021confound}. While disentanglement methods have shown effectiveness in improving fairness, they often come with a performance-fairness trade-off. By removing or suppressing information related to sensitive attributes, the model can inadvertently remove clinically relevant information, potentially harming overall performance. Sensitive attributes are often clinically relevant; for example, certain diseases have different prevalence or manifestation across age or sex groups. 

In this paper, we propose \textbf{Fair} \textbf{RE}-fusion \textbf{A}fter \textbf{D}isentanglement (\textbf{FairREAD}), an innovative approach that aims to mitigate unfairness without sacrificing model performance. FairREAD is constructed upon a simple yet effective core idea: to make use of information from both the input image and the sensitive demographic attributes. By first removing demographic attributes-related information from the image representation and then \rev{fusing} them back in, we could make use of the demographic attributes in diagnosis while removing potential spurious correlations hidden in the image representation. To summarize, our contributions are threefold:
\begin{itemize}
    \item We employ a fair image encoder that leverages orthogonality constraints and adversarial training to decouple sensitive demographic attributes from learned representations, mitigating unintended biases.
    \item We propose a re-fusion mechanism that that incorporates demographic information in a controlled manner to balance fairness and performance.
    \item We introduce a subgroup-specific threshold adjustment strategy, enabling the mitigation of performance disparities across demographic groups while maintaining diagnostic accuracy.
\end{itemize}
By integrating these innovations, FairREAD establishes a comprehensive strategy to balance fairness and accuracy.
Through extensive evaluation on the CheXpert dataset, a standard benchmark for chest X-ray analysis, we demonstrate significant advancement in fairness metrics without compromising diagnostic performance.

\section{Methods}

\begin{figure*}
    \centering
    \includegraphics[width=0.95\linewidth]{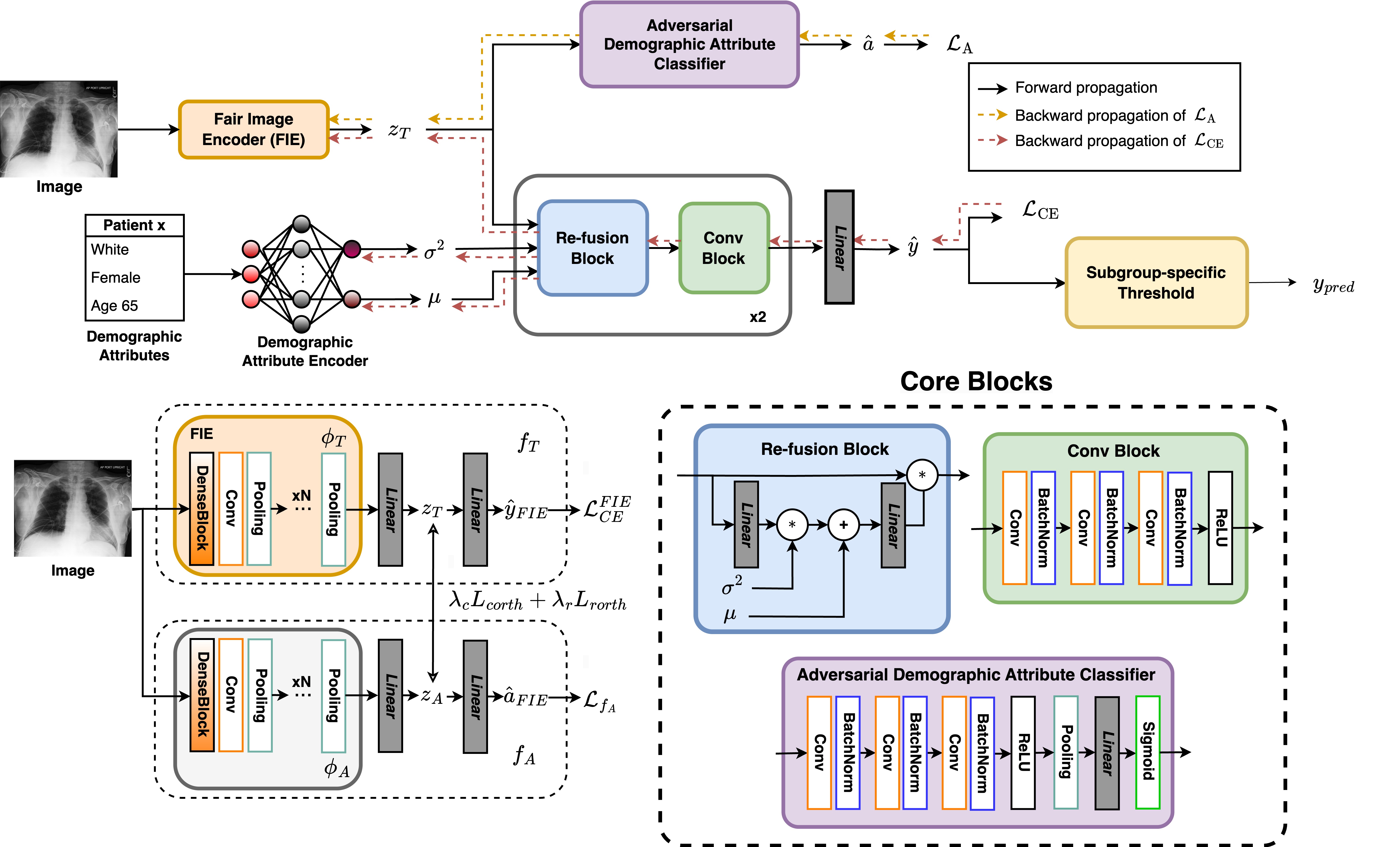}
    \caption{Model architecture of FairREAD. The input chest X-ray image is first encoded by the Fair Image Encoder (FIE) into a demographic attribute-invariant representation $\zt$. An adversarial demographic attribute classifier is trained to ensure demographic attributes are not recoverable from $\zt$. Then $\zt$ is passed into a re-fusion module including re-fusion blocks and convolution blocks that integrates demographic information through rescaling in latent space. The output logit of the entire pipeline is then passed through a subgroup-specific threshold to derive the classification result.}
    \label{fig:model-arch}
\end{figure*}

The overall architecture of our proposed model, FairREAD, is illustrated in Figure \ref{fig:model-arch}. There are three key features included in our model: fair image encoder (\ref{subsec:fair-encoder}), re-fusion mechanism (\ref{subsec:refusion}), and subgroup-specific classification threshold (\ref{subsec:threshold}). The aim of our model is to make use of information from both the input image and the demographic attributes, and at the same time mitigate unfairness of the classification output. 

In this section, we also explain the detailed implementation of FairREAD (\ref{subsec:implementation}), and the dataset (\ref{subsec:dataset}), baseline methods (\ref{subsec:baselines}) and metrics (\ref{subsec:metrics}) we use in our experiments.

\subsection{Fair image encoder construction}\label{subsec:fair-encoder}

The input image is first passed into a fair image encoder, from which we aim to obtain a fair representation of the image that contains no information related to the demographic attributes. To construct the fair image encoder, we design a novel training scheme for the encoder such that our target representation is independent of the sensitive representation, where combined orthogonality and adversarial training are leveraged. 

Consider a binary image classification problem with input images $x\in \mathcal{X}$, and ground truth labels $y\in \{0, 1\}$. Suppose each input sample has $\da$ demographic attributes given as a column vector $a\in \{0, 1\}^\da$.\footnote{For simplicity, we assume that each demographic attribute is binary. Implementation details regarding discretization of demographic attributes will be discussed in Section \ref{subsec:dataset}.} Our goal is to train an image-to-target classifier $\ft:\mathcal{X}\rightarrow\{0, 1\}$ with $n$ training samples. Suppose that $\ft$ is composed of two parts: an image encoder $\phit:\mathcal{X}\rightarrow\RR^\dz$ and a classification head $\headt:\RR^\dz\rightarrow\{0, 1\}$, where the output of $\phit$, defined as latent vector $\zt\in\RR^\dz$, is the input to $\headt$. In order to mitigate unfairness in $\ft$, we also train two auxiliary models: an image-to-demographic-attribute classifier $\fa:\mathcal{X}\rightarrow\RR^\da$, and an adversarial latent-vector-to-demographic-attribute classifier $\fadv:\RR^\dz\rightarrow\RR^\da$. We denote the image encoder part of $\fa$ as $\phia:\mathcal{X}\rightarrow\RR^\dz$, and the classification head of $\fa$ as $\heada:\RR^\dz\rightarrow\RR^\da$. Note that the dimension of the latent space vector in $\fa$ is the same as $\ft$. Let $\Zt\in\RR^{n\times\dz}$ be a matrix of latent space vectors given by $\phit$ with all training samples, where each row $\Ztrow\in\RR^\dz$ corresponds to one training sample, and each column $\Ztcol\in\RR^n$ corresponds to one latent feature. Similarly, let $\Za\in\RR^{n\times\dz}$ be a latent space matrix given by $\phia$, where each row is denoted as $\Zarow\in\RR^\dz$ and each column is $\Zacol\in\RR^n$. With the definitions above, we construct three loss functions, described in the following.

\noindent\textbf{Column space orthogonality loss:}. The goal of the column space orthogonality loss is to ensure that the target latent vector $\zt$ has minimal projection onto the sensitive space, meaning that the information about the target is maximally independent of the demographic attributes. Concretely, we first construct a low-rank space $\Tilde{\mathcal{S}_A}$ using Singular Value Decomposition (SVD) on $\Za$:
\begin{equation}
    \Za = U_A\Sigma_AV_A
\end{equation}
where $U_A\in\RR^{n\times n}$, $V_A\in\RR^{\dz\times\dz}$, and $\Sigma_A\in\RR^{n\times\dz}$ contains the left singular vectors, right singular vectors, and singular values of $\Za$, respectively. The $k$ most important left singular vectors in $U_A$ are denoted as $\Tilde{S_A}\in\RR^{n\times k}$. We then compute the column space orthogonality loss:
\begin{equation}
    L_{corth} = \sum\limits_{i=1}^n\frac{||\Tilde{S_A^T}\Ztcol||_2^2}{||\Ztcol||_2^2}
\end{equation}
which represents the projection of $\Zt$ to $\Tilde{S_A}$. 

\noindent\textbf{Row space orthogonality loss:}. The goal of the row space orthogonality loss is to make each feature dimension of the target representation orthogonal to the corresponding feature dimension in the sensitive representation. Concretely, we achieve this by minimizing the mean value of covariance between each row $\Ztrow$ in $\Zt$ and each row $\Zarow$ in $\Za$:
\begin{equation}
    L_{rorth} = \frac{1}{d^2}\sum\limits_{i=1}^\dz\sum\limits_{j=1}^\dz[(\Ztrow - \mu_T)(\Zarow - \mu_A)^T]^2
\end{equation}
where $\mu_T, \mu_A\in\RR^\dz$ represents the row-wise mean vector of $\Zt$ and $\Za$, respectively.

\noindent\textbf{Adversarial loss}. The goal of the adversarial loss is to ensure that the demographic attributes are not recognizable from the target latent vector $\zt$. To achieve this, we pass $\zt$ into the adversarial classifier $f_{adv}$, which is trained to predict the demographic attribute vector $a$. We then use the negative cross-entropy loss of $f_{adv}$ to train the target model:
\begin{align}
    L_{adv} &= -\text{Cross-Entropy}(a, \hat{a}) \\
    &= -\sum\limits_{i=1}^{\da}-a_i\log\hat{a_i} + (1 -a_i)\log(1 - \hat{a_i})
\end{align}
where $\hat{a}$ is the predicted demographic attributes given by $f_{adv}$.

These loss functions are applied in various stages of the training process to ensure no information related to demographic attributes in the output of the fair image encoder. Details of the training process will be discussed in Section \ref{subsec:implementation}.

\subsection{Re-fusion mechanism}\label{subsec:refusion}

In order to make use of information from the demographic attributes, we propose a novel re-fusion mechanism, which aims to fuse the fair image representation and the demographic attributes.

To begin with, the demographic attributes are encoded using a Multi-Layer Perceptron (MLP) \rev{with} two outputs. \rev{We denote the outputs as $\mu$ and $\sigma^2$, which indicates the mean and variance for feature rescaling.} Next, in the re-fusion block, the fair image representation \rev{$\zt\in\RR^\dz$} is projected to a lower dimension \rev{$d_{hidden}$}, rescaled using $\mu$ and $\sigma^2$, and projected back. \rev{The value of the lower dimension $d_{hidden}$ is a hyperparameter.} Then, the rescaled representation and the fair representation are fused via element-wise multiplication. Concretely, given an input image $x$ and a sensitive attribute vector $a$, the fused representation $z_{fused}$ is derived as follows.
\begin{align}
    & \zt = \text{FairImageEncoder}(x) \\
    & \mu, \sigma^2 = \text{MLP}(a) \\
    & z_{fused} = \text{Re-fusion}(\zt, \mu, \sigma^2)
\end{align}
where
\begin{align}
    \text{Re-fusion}(\zt, \mu, \sigma^2) = \zt * \text{Proj}^{-1}(\sigma^2 * \text{Proj}(\zt) + \mu)
\end{align}
where $*$ represents element-wise product, $\text{Proj}, \text{Proj}^{-1}$ are both single fully-connected layers. The fused representation $z_{fused}$ is then passed into a convolution block. This sequence of a re-fusion block followed by a convolution block is repeated \rev{for $N$ times, where $N$ is a hyperparameter}. Finally, the output is passed through a fully-connected layer that generates the predicted logit $\hat{y}$.

\subsection{Subgroup-specific classification threshold}\label{subsec:threshold}

After a predicted logit $\hat{y}$ has been generated as described above, we apply subgroup-specific classification threshold to further improve fairness across each demographic subgroup. Specifically, for each subgroup $g$, we determine a unique classification threshold $\theta_g$ by minimizing the absolute difference between the True Positive Rate (TPR) and the True Negative Rate (TNR) within that group:

\begin{align}
    \theta_g &= \arg\min_{s} |\text{TPR}_{g, s}-\text{TNR}_{g, s}|
\end{align}
where $\text{TPR}_{g, s}$ and $\text{TNR}_{g, s}$ denote the True Positive Rate and True Negative Rate for group $g$ given a classification threshold value of $s$. We refer to this threshold selection strategy as the \textbf{Min-gap} strategy.  After $\theta_g$ is determined for each demographic subgroup $g$, we apply it to derive the final classification result $y_{pred}\in\{0, 1\}$:

\begin{align}
    y_{pred} = \mathbb{I}(\hat{y} > \theta_{g_x})
\end{align}
where $\mathbb{I}$ is the indicator function, $g_x$ is the demographic subgroup of the input $x$ that generated the output logit $\hat{y}$. \rev{Please note that the threshold selection of the Min-gap strategy is done on the training set, where label imbalance and subgroup imbalance are present. Full details is described in Section \ref{subsec:dataset}.}

\subsection{Implementation details}\label{subsec:implementation}

The training process of FairREAD is taken out in two stages. In the first stage, we train the fair image encoder using column and row orthogonality loss described in Section \ref{subsec:fair-encoder}. Concretely, we first train the image-to-demographic-attribute classifier $\fa$ using the cross-entropy loss function:

\begin{align}
    \L_{\fa} = \text{Cross-Entropy}(a, \rev{\hat{a}_{FIE}})
\end{align}
where \rev{$a$ and $\hat{a}_{FIE}$} represents the ground-truth and predicted demographic attributes, respectively. Next, we train the image-to-target classifier $\ft$ using the following loss function:

\begin{align}
    \L_{\ft} &= \mathcal{L}_{CE}^{FIE} + \lambda_{c}\L_{corth} + \lambda_{r}\L_{rorth} \\
    &= \text{Cross-Entropy}(y, \rev{\hat{y}_{FIE}}) + \lambda_{c}\L_{corth} + \lambda_{r}\L_{rorth}
\end{align}
where \rev{$y$ and $\hat{y}_{FIE}$ represents the ground truth and the predicted target label, respectively.} $\lambda_{c}$ and $\lambda_{r}$ are hyperparameters controlling the weight of column- and row-orthogonality loss. After the first stage training is complete, we use the encoder part ($\phit$) of $f_T$ as the fair image encoder and plug it into the FairREAD architecture shown in Figure \ref{fig:model-arch}.

In the second stage, we train the FairREAD model as a whole. We use a cross-entropy loss:

\begin{align}
    \L_{CE} = \text{Cross-Entropy}(y, \hat{y})
\end{align}

Additionally, in order to protect the fair image encoder from being perturbed during the second stage training, we apply the adversarial loss described in Section \ref{subsec:fair-encoder}. Concretely, we iteratively train an adversarial demographic attribute classifier $\fadv$ and the fair image encoder. We use a cross-entropy loss $\L_A$ to optimize $\fadv$:

\begin{align}
    \L_{A} = \text{Cross-Entropy}(a, \hat{a})
\end{align}

We then subtract $\L_{A}$ from $\L_{CE}$ described above to obtain the loss function of FairREAD during the second stage of training:

\begin{align}
    \L_{\text{FairREAD}} = \L_{\text{CE}} - \rev{\alpha_{adv}} * \L_{A}
\end{align}

where \rev{$\alpha_{adv}$} is a hyperparameter that controls the degree of adversarial training. During back-propagation, the fair image encoder receives a combined gradient from both $\L_{CE}$ and $-\L_{A}$, which guides it to maximize the loss of $A$, while ensuring the classification performance of the entire FairREAD model. Since other parts of the FairREAD model do not contribute to $\L_A$, they do not receive gradient from $\L_A$ and are equivalent to being  optimized using $\L_{\text{CE}}$ only.

\subsection{Dataset}\label{subsec:dataset}

\begin{table*}[h]
\centering
\footnotesize
\caption{Number of Samples, Ratios, and Positive Rates by Each Sensitive Attribute}
\label{tab:dataset}
\setlength{\tabcolsep}{3pt} %
\begin{subtable}[t]{0.4\textwidth}
\centering
\scriptsize
\caption{Number of Samples and Ratios}
\label{tab:samples_ratios}
\begin{tabular}{lcccccc}
\toprule
 & \multicolumn{2}{c}{\textbf{Gender}} & \multicolumn{2}{c}{\textbf{Race}} & \multicolumn{2}{c}{\textbf{Age}} \\
\cmidrule(lr){2-3}  \cmidrule(lr){4-5}  \cmidrule(lr){6-7}
\textbf{Pathology} & \textbf{Male} & \textbf{Female} & \textbf{White} & \textbf{Non-White} & \textbf{$\geq60$} & \textbf{$<60$} \\
\midrule
\textbf{Cardiomegaly} &
\begin{tabular}{@{}c@{}}88,458\\(75.28\%)\end{tabular} &
\begin{tabular}{@{}c@{}}29,041\\(24.72\%)\end{tabular} &
\begin{tabular}{@{}c@{}}75,866\\(64.57\%)\end{tabular} &
\begin{tabular}{@{}c@{}}41,633\\(35.43\%)\end{tabular} &
\begin{tabular}{@{}c@{}}70,184\\(59.73\%)\end{tabular} &
\begin{tabular}{@{}c@{}}47,315\\(40.27\%)\end{tabular} \\
\midrule
\textbf{Pleural Effusion} &
\begin{tabular}{@{}c@{}}92,981\\(65.95\%)\end{tabular} &
\begin{tabular}{@{}c@{}}48,009\\(34.05\%)\end{tabular} &
\begin{tabular}{@{}c@{}}82,192\\(58.30\%)\end{tabular} &
\begin{tabular}{@{}c@{}}58,798\\(41.70\%)\end{tabular} &
\begin{tabular}{@{}c@{}}87,191\\(61.84\%)\end{tabular} &
\begin{tabular}{@{}c@{}}53,799\\(38.16\%)\end{tabular} \\
\midrule
\textbf{Fracture} &
\begin{tabular}{@{}c@{}}62,046\\(80.66\%)\end{tabular} &
\begin{tabular}{@{}c@{}}14,876\\(19.34\%)\end{tabular} &
\begin{tabular}{@{}c@{}}54,407\\(70.73\%)\end{tabular} &
\begin{tabular}{@{}c@{}}22,515\\(29.27\%)\end{tabular} &
\begin{tabular}{@{}c@{}}55,766\\(72.50\%)\end{tabular} &
\begin{tabular}{@{}c@{}}21,156\\(27.50\%)\end{tabular} \\
\midrule
\textbf{Pneumonia} &
\begin{tabular}{@{}c@{}}55,909\\(78.92\%)\end{tabular} &
\begin{tabular}{@{}c@{}}14,938\\(21.08\%)\end{tabular} &
\begin{tabular}{@{}c@{}}51,625\\(72.87\%)\end{tabular} &
\begin{tabular}{@{}c@{}}19,222\\(27.13\%)\end{tabular} &
\begin{tabular}{@{}c@{}}56,056\\(79.12\%)\end{tabular} &
\begin{tabular}{@{}c@{}}14,791\\(20.88\%)\end{tabular} \\
\bottomrule
\end{tabular}
\end{subtable}
\hfill
\begin{subtable}[t]{0.4\textwidth}
\centering
\scriptsize
\caption{Positive Rates (\%)}
\label{tab:positive_rates}
\begin{tabular}{lcccccc}
\toprule
 & \multicolumn{2}{c}{\textbf{Gender}} & \multicolumn{2}{c}{\textbf{Race}} & \multicolumn{2}{c}{\textbf{Age}} \\
\cmidrule(lr){2-3}  \cmidrule(lr){4-5}  \cmidrule(lr){6-7}
\textbf{Pathology} & \textbf{Male} & \textbf{Female} & \textbf{White} & \textbf{Non-White} & \textbf{$\geq60$} & \textbf{$<60$} \\
\midrule
\textbf{Cardiomegaly} &
14.86\% & 30.12\% & 13.46\% & 28.05\% & 24.31\% & 10.21\% \\
\midrule
\textbf{Pleural Effusion} &
48.15\% & 64.85\% & 42.64\% & 69.50\% & 60.12\% & 43.66\% \\
\midrule
\textbf{Fracture} &
9.54\% & 20.83\% & 8.32\% & 19.94\% & 8.87\% & 19.24\% \\
\midrule
\textbf{Pneumonia} &
6.33\% & 16.53\% & 5.29\% & 17.03\% & 6.26\% & 16.88\% \\
\bottomrule
\end{tabular}
\end{subtable}
\end{table*}

\begin{figure*}[tbh!]
    \centering
    \includegraphics[width=0.8\linewidth]{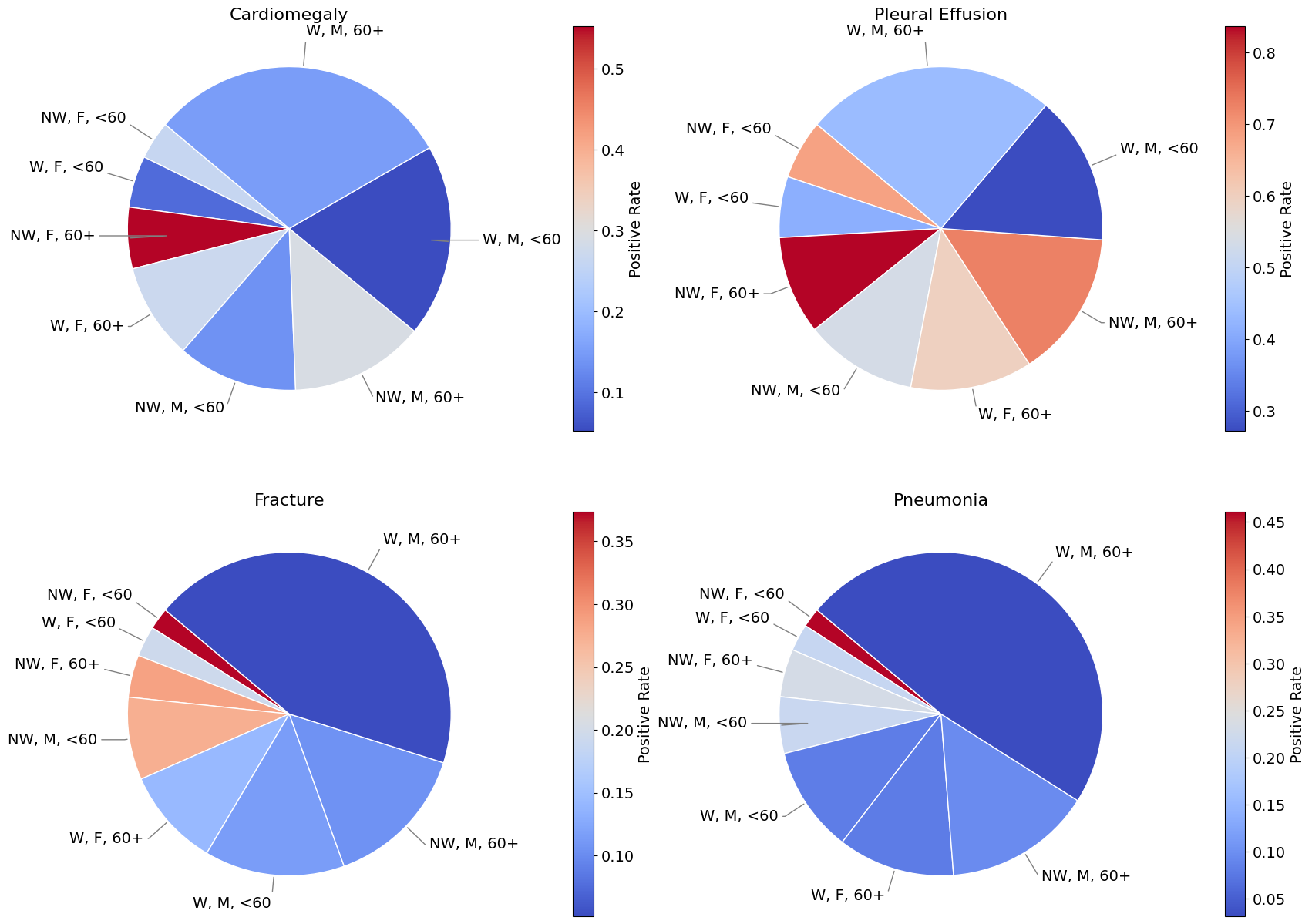}
    \caption{Distribution of demographic subgroups in the processed CheXpert dataset. ``NW" and ``W" denotes ``Non-White" and ``White"; ``M" and ``F" represent ``Male" and ``Female," respectively; ``$<60$" indicates patients younger than 60 years, while ``60+" indicates patients aged 60 years or older.}
    \label{fig:dataset}
\end{figure*}

We conduct our experiments on CheXpert \citep{CheXpert}, a large public dataset containing 224,316 chest X-ray images of 65,240 patients. We focus on four \rev{pathologies} in CheXpert: Cardiomegaly, Pleural Effusion, Pneumonia, and Fracture. We consider three demographic attributes: gender, age, and primary race. For simplicity, we discretize each demographic attribute into binary values. For age, we classify all patients as less than 60 years and 60 years or older; For race, we classify all patients as White and Non-white. For gender, since the dataset only contains two gender values (Male and Female), we use these labels directly. In order to better manifest the effectiveness of our method in unfairness mitigation, we magnify subgroup disparity of the dataset by subsampling CheXpert to create larger disparities in positive rate and number of samples across demographic subgroups. Concretely, for Cardiomegaly and Pleural Effusion, we subsample the training set to create a positive rate disparity of around 16\%. For Pneumonia and Fracture, we only use a positive rate disparity of around 10\% since a larger value would result in the number of samples within some subgroups being too small. Details of subgroup distribution in the dataset is shown in Table \ref{tab:dataset} and Figure \ref{fig:dataset}.

In order to compare the performance of our model and the baseline models, we hold out 10\% of the data as our test set, and conduct a five-fold cross-validation on the other 90\% of the data. \rev{For each fold, we train the model on the training set for ten epochs. After each epoch, we test the model on the validation set, and do early stopping if the validation loss does not decrease for three consecutive epochs. Then, we fit a decision threshold on the training set with the min-gap strategy introduced in Section~\ref{subsec:threshold}. All hyperparameters in FairREAD are decided using the average performance on validation sets across all five folds. More details regarding hyperparameter searches are discussed in \ref{app:hyperparamter}. We then test the model's performance on the test set, and take the average and standard deviation across all five folds.} The data split is fixed between the two stages of training to avoid data contamination. 

\subsection{Baseline methods}\label{subsec:baselines}

In order to manifest the effectiveness of our method, we compare our model to the following baselines:

\textbf{Empirical Risk Minimization (ERM)}. ERM stands for the vanilla training process that only optimizes against the classification cross-entropy loss $\L_{CE}$, and without any measures to improve fairness. 

\rev{\textbf{Fairness via Column-Row
space Orthogonality (FCRO)}} \citep{FairOrth}. \rev{FCRO} first trains a demographic attribute classifcation model, and then trains a target label classification label with a combination of minimizing the classification loss and maximizing row-space and column-space orthogonality between the hidden space of both models.

\textbf{Learning Not To Learn (LNTL)} \citep{LNTL, LNTLmi}. LNTL trains a separate adversarial model to recognize the demographic attributes given the \rev{image representation} calculated with an encoder. Besides, a classification head is trained to predict the target label given the \rev{image representation}. The encoder is trained with a combination of minimizing the loss of the classification head and maximizing the loss of the adversarial model. 

\textbf{Adaptive Batch Normalization (FairAdaBN)} \citep{FairAdaBN}. FairAdaBN substitutes all the batch normalization layers within the model with a subgroup-specific batch normalization, where each subgroup is assigned to a separate set of batch normalization parameters.

\textbf{Adversarial Learning (AL)} \citep{adv}. AL trains a separate adversarial model trained to recognize the demographic attributes given the output logit of the original classification model. The classification model is trained with a combination of minimizing the classification loss and maximizing the loss of the adversarial model.

To ensure fair comparison, we apply subgroup-specific threshold selection introduced in Section \ref{subsec:threshold} for all baseline methods. Concretely, we fit the optimal classification threshold using the Min-gap strategy after the training process.

\subsection{Evaluation metrics}\label{subsec:metrics}

We evaluate all the methods above for both performance and fairness. In terms of performance, we measure classification accuracy and AUC. In terms of fairness, following previous works \citep{FairOrth, FairAdaBN, chestxray-fairness-zhang}, we use subgroup disparity with respect to Equal Odds ($\Delta_{ED}$), and AUC ($\Delta_{AUC}$).  Given predicted labels $\hat{Y}$, ground truth labels $Y$, and corresponding subgroup of each sample $A$, the fairness metrics are defined as follows:
\begin{align}
    & \Delta_{AUC} = \max\limits_{a_0, a_1 \in \mathcal{A}} |AUC_{a_0} - AUC_{a_1}| \\
    & \Delta_{ED} = \max\limits_{\substack{y \in \{0, 1\} \\ a_0, a_1 \in \mathcal{A}}} |P(\hat{Y}=0|Y=y, A=a_0) - P(\hat{Y}=y|Y=0, A=a_1)|
\end{align}
where $\mathcal{A}$ is the set of all subgroups, $AUC_a$ is the AUC of the model on all samples belonging to subgroup $a$.

Additionally, we use $FATE_{EO}$ and $FATE_{AUC}$ \citep{FairAdaBN} as comprehensive measures of both performance and fairness. These metrics measure the trade-off quality between performance and fairness, defined as follows:
\begin{align}
    & FATE_{EO} = \frac{Acc - Acc_{ERM}}{Acc_{ERM}} - \frac{\Delta_{EO} - \Delta_{EO, ERM}}{\Delta_{EO, ERM}} \\
    & FATE_{AUC} = \frac{AUC - AUC_{ERM}}{AUC_{ERM}} - \frac{\Delta_{EO} - \Delta_{AUC, ERM}}{\Delta_{AUC, ERM}}
\end{align}
where $Acc_{ERM}, \Delta_{EO, ERM}, AUC_{ERM}, \Delta_{AUC, ERM}$ means the accuracy, $\Delta_{EO}$, AUC, and $\Delta_{AUC}$ of the ERM model trained using the same dataset as the measured model, respectively.

\section{Experimental results}
\subsection{Model performance}

\begin{table*}[tbh!]
\centering
\footnotesize
\caption{Results for all diseases. Best performance is marked in bold, second best performance is underlined.}\label{tab:results-main}
\begin{tabular}{l l c c c c c c}
\toprule
\rev{\textbf{Pathology}} & \textbf{Method} & Accuracy($\uparrow$) & AUC($\uparrow$) & $\Delta_{AUC}$($\downarrow$) & $\Delta_{EO}$($\downarrow$) & $FATE_{EO}$($\uparrow$) & $FATE_{AUC}$($\uparrow$) \\
\midrule
 & ERM &
\textbf{0.815 $\pm$ 0.017} &
0.853 $\pm$ 0.007 &
0.075 $\pm$ 0.007 &
0.176 $\pm$ 0.014 &
0.000 &
0.000 \\
\cmidrule{2-8}

 & FCRO &
0.795 $\pm$ 0.003 &
0.842 $\pm$ 0.007 &
0.082 $\pm$ 0.022 &
\underline{0.133 $\pm$ 0.021} &
\underline{0.220} &
-0.116 \\
\cmidrule{2-8}

Cardiomegaly & LNTL &
0.749 $\pm$ 0.029 &
0.836 $\pm$ 0.005 &
0.104 $\pm$ 0.011 &
0.152 $\pm$ 0.058 &
0.055 &
-0.409 \\
\cmidrule{2-8}

 & FairAdaBN &
0.795 $\pm$ 0.009 &
\underline{0.877 $\pm$ 0.001} &
0.075 $\pm$ 0.010 &
0.140 $\pm$ 0.023 &
0.179 &
0.021 \\
\cmidrule{2-8}

 & AL &
\underline{0.806 $\pm$ 0.026} &
0.873 $\pm$ 0.003 &
\textbf{0.073 $\pm$ 0.002} &
0.143 $\pm$ 0.024 &
0.176 &
\underline{0.047} \\

\cmidrule{2-8}

 & \textbf{FairREAD (ours)} &
0.781 $\pm$ 0.004 &
\textbf{0.880 $\pm$ 0.003} &
\underline{0.073 $\pm$ 0.007} &
\textbf{0.113 $\pm$ 0.006} &
\textbf{0.318} &
\textbf{0.050} \\
\midrule
 & ERM &
0.796 $\pm$ 0.006 &
0.888 $\pm$ 0.002 &
0.047 $\pm$ 0.004 &
0.079 $\pm$ 0.012 &
0.000 &
0.000 \\
\cmidrule{2-8}

 & FCRO &
0.793 $\pm$ 0.002 &
0.880 $\pm$ 0.001 &
\underline{0.040 $\pm$ 0.002} &
\textbf{0.071 $\pm$ 0.010} &
\textbf{0.101} &
\underline{0.149} \\
\cmidrule{2-8}

Pleural Effusion & LNTL &
0.771 $\pm$ 0.005 &
0.866 $\pm$ 0.004 &
0.046 $\pm$ 0.005 &
0.088 $\pm$ 0.012 &
-0.143 &
0.004 \\
\cmidrule{2-8}

 & FairAdaBN &
\underline{0.801 $\pm$ 0.004} &
\underline{0.896 $\pm$ 0.001} &
0.048 $\pm$ 0.009 &
0.081 $\pm$ 0.006 &
-0.021 &
0.006 \\
\cmidrule{2-8}

 & AL &
0.796 $\pm$ 0.005 &
0.889 $\pm$ 0.002 &
0.041 $\pm$ 0.002 &
0.080 $\pm$ 0.012 &
-0.011 &
0.137 \\

\cmidrule{2-8}

 & \textbf{FairREAD (ours)} &
\textbf{0.807 $\pm$ 0.002} &
\textbf{0.901 $\pm$ 0.002} &
\textbf{0.038 $\pm$ 0.003} &
\underline{0.077 $\pm$ 0.011} &
\underline{0.045} &
\textbf{0.218} \\
\midrule
 & ERM &
0.691 $\pm$ 0.012 &
0.778 $\pm$ 0.004 &
0.096 $\pm$ 0.022 &
0.197 $\pm$ 0.038 &
0.000 &
0.000 \\
\cmidrule{2-8}

 & FCRO &
\textbf{0.780 $\pm$ 0.040} &
0.715 $\pm$ 0.021 &
0.118 $\pm$ 0.025 &
0.193 $\pm$ 0.024 &
\underline{0.149} &
-0.307 \\
\cmidrule{2-8}

Pneumonia & LNTL &
0.652 $\pm$ 0.026 &
0.740 $\pm$ 0.017 &
0.107 $\pm$ 0.017 &
0.179 $\pm$ 0.068 &
0.033 &
-0.156 \\
\cmidrule{2-8}

 & FairAdaBN &
0.629 $\pm$ 0.042 &
\underline{0.783 $\pm$ 0.014} &
0.100 $\pm$ 0.015 &
\underline{0.172 $\pm$ 0.036} &
0.036 &
-0.033 \\
\cmidrule{2-8}

 & AL &
\underline{0.770 $\pm$ 0.035} &
0.764 $\pm$ 0.010 &
\textbf{0.084 $\pm$ 0.008} &
0.218 $\pm$ 0.036 &
0.008 &
\underline{0.110} \\

\cmidrule{2-8}

 & \textbf{FairREAD (ours)} &
0.677 $\pm$ 0.022 &
\textbf{0.804 $\pm$ 0.008} &
\underline{0.085 $\pm$ 0.026} &
\textbf{0.146 $\pm$ 0.024} &
\textbf{0.237} &
\textbf{0.155} \\
\midrule
 & ERM &
0.674 $\pm$ 0.020 &
\underline{0.757 $\pm$ 0.006} &
0.164 $\pm$ 0.018 &
0.220 $\pm$ 0.020 &
0.000 &
0.000 \\
\cmidrule{2-8}

& FCRO &
\textbf{0.759 $\pm$ 0.008} &
0.717 $\pm$ 0.009 &
0.129 $\pm$ 0.015 &
0.252 $\pm$ 0.038 &
-0.066 &
\underline{0.229} \\
\cmidrule{2-8}

Fracture & LNTL &
0.597 $\pm$ 0.063 &
0.721 $\pm$ 0.014 &
\underline{0.118 $\pm$ 0.024} &
0.197 $\pm$ 0.035 &
-0.009 &
\underline{0.229} \\
\cmidrule{2-8}

 & FairAdaBN &
0.640 $\pm$ 0.015 &
0.749 $\pm$ 0.008 &
0.141 $\pm$ 0.013 &
\textbf{0.172 $\pm$ 0.017} &
\underline{0.166} &
0.125 \\
\cmidrule{2-8}

 & AL &
\underline{0.728 $\pm$ 0.061} &
0.756 $\pm$ 0.008 &
0.146 $\pm$ 0.021 &
0.252 $\pm$ 0.024 &
-0.066 &
0.107 \\
\cmidrule{2-8}

 & \textbf{FairREAD (ours)} &
0.678 $\pm$ 0.035 &
\textbf{0.770 $\pm$ 0.009} &
\textbf{0.107 $\pm$ 0.009} &
\underline{0.181 $\pm$ 0.024} &
\textbf{0.183} &
\textbf{0.364} \\

\midrule

 & ERM &
0.744 $\pm$ 0.007 &
0.819 $\pm$ 0.002 &
0.096 $\pm$ 0.007 &
0.168 $\pm$ 0.012 &
0.000 &
0.000 \\
\cmidrule{2-8}

\textbf{} & FCRO &
\textbf{0.782 $\pm$ 0.010 }&
0.788 $\pm$ 0.006 &
0.092 $\pm$ 0.009 &
0.162 $\pm$ 0.013 &
\underline{0.112} &
-0.029 \\
\cmidrule{2-8}

\textbf{Average}& LNTL &
0.692 $\pm$ 0.019 &
0.791 $\pm$ 0.006 &
0.094 $\pm$ 0.008 &
0.154 $\pm$ 0.024 &
-0.016 &
-0.083 \\
\cmidrule{2-8}

\textbf{} & FairAdaBN &
0.716 $\pm$ 0.011 &
\underline{0.826 $\pm$ 0.004} &
0.091 $\pm$ 0.006 &
\underline{0.141 $\pm$ 0.012} &
0.090 &
0.030 \\
\cmidrule{2-8}

\textbf{} & AL &
\underline{0.775 $\pm$ 0.019} &
0.820 $\pm$ 0.003 &
\underline{0.086 $\pm$ 0.006} &
0.173 $\pm$ 0.013 &
0.027 &
\underline{0.100} \\

\cmidrule{2-8}

\textbf{} & \textbf{FairREAD (ours)} &
0.736 $\pm$ 0.010 &
\textbf{0.839 $\pm$ 0.003} &
\textbf{0.076 $\pm$ 0.007} &
\textbf{0.129 $\pm$ 0.009} &
\textbf{0.196} &
\textbf{0.197} \\
\bottomrule
\end{tabular}
\end{table*}

\begin{figure*}[tbh]
    \centering
\includegraphics[width=\linewidth]{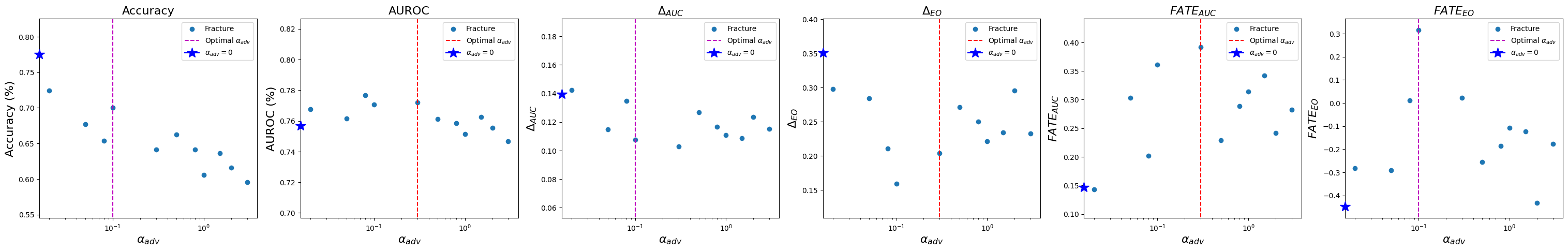}
    \caption{\rev{Performance of FairREAD on classifying patients with Fracture using different values of $\alpha_{adv}$. Optimal value of $\alpha_{adv}$ in terms of ${FATE}_{EO}$ is marked in \rev{purple} dashed lines in the figures for Accuracy, $\Delta_{EO}$, and ${FATE}_{EO}$. Optimal value of $\alpha_{adv}$ in terms of ${FATE}_{AUC}$ is marked in orange dashed lines in the figures for AUC, $\Delta_{AUC}$, and ${FATE}_{AUC}$. The value in each plot when $\alpha_{adv} = 0$ is marked with a blue star.}}
    \label{fig:ablation-alpha}
\end{figure*}

\begin{figure*}[htbp]
    \centering
    \begin{subfigure}[b]{0.48\linewidth}
        \centering
        \includegraphics[width=\linewidth]{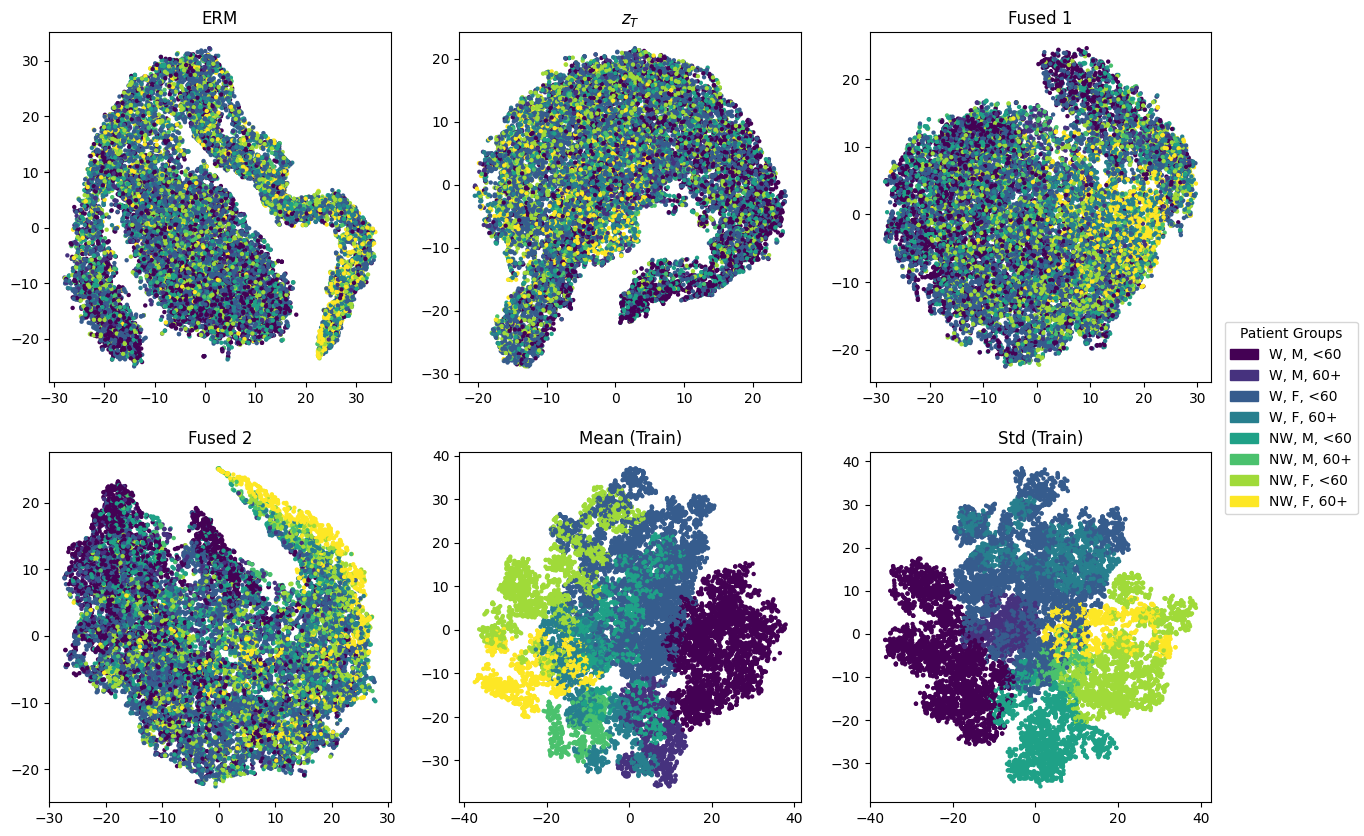}
        \caption{Embedding space colored by demographic subgroups}
        \label{fig:emb-subgroup}
    \end{subfigure}
    \hfill
    \begin{subfigure}[b]{0.48\linewidth}
        \centering
        \includegraphics[width=\linewidth]{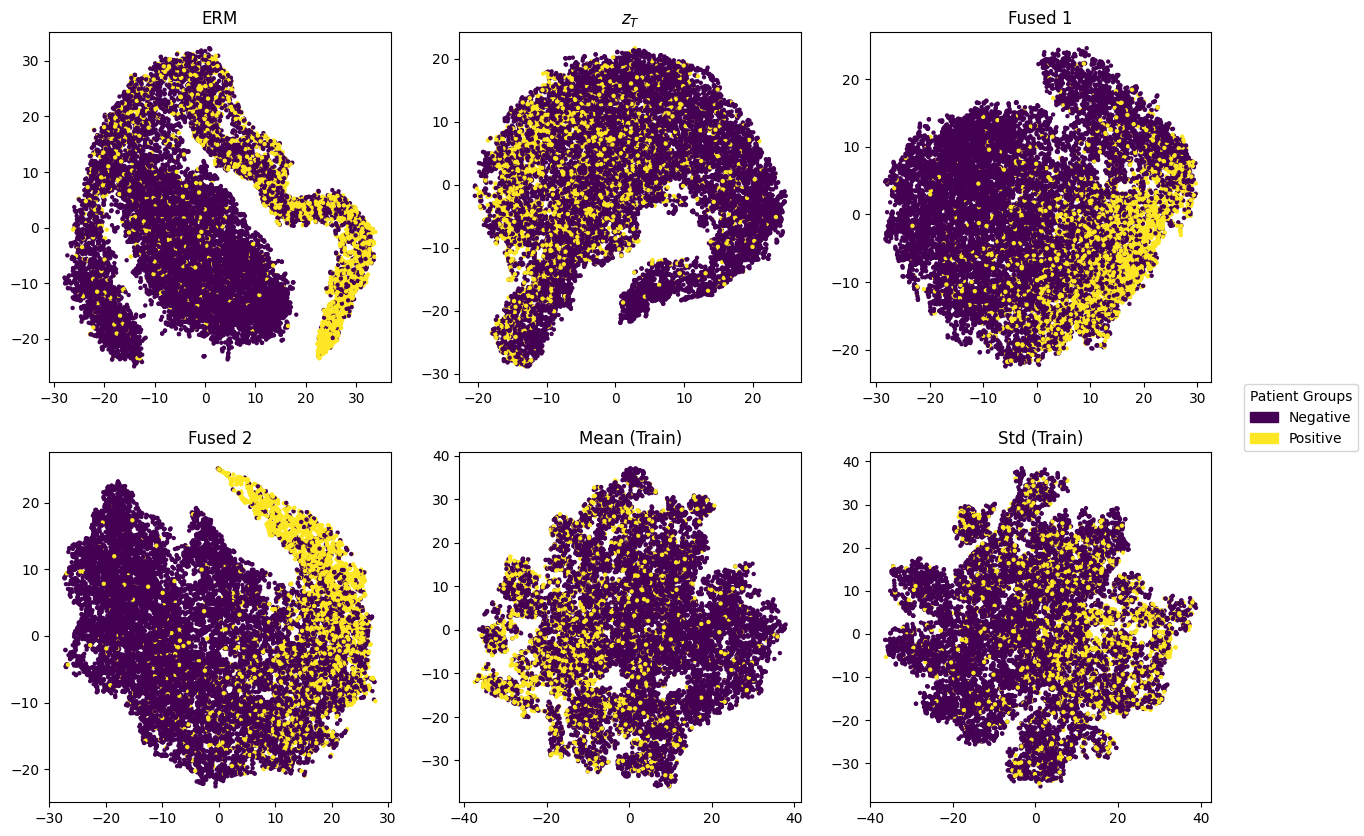}
        \caption{Embedding space colored by target labels (Cardiomegaly)}
        \label{fig:emb-target}
    \end{subfigure}
    \caption{t-SNE visualization of embedding space in FairREAD model for Cardiomegaly classification (best viewed in color). ``ERM" represents the embedding space of the ERM model, ``$\zt$", ``Fused 1", ``Fused 2" corresponds to the embedding space after fair image encoder, the first re-fusion \& convolution block, and the second re-fusion \& convolution block, respectively. ``Mean (Train)" and ``Std (Train)" corresponds to the outputs of the demographic attribute encoder ($\mu$ and $\sigma^2$). For best visualization effect, we show the ``Mean" and ``Std" embedding space during training, when dropout layers in the model are activate to add some noise between samples in the same demographic subgroup. Other embedding spaces are visualized when the model is in evaluation mode.}
    \label{fig:emb-combined}
\end{figure*}

The results of our experiments are listed in Table \ref{tab:results-main}. We can observe the following:

\textbf{On average, FairREAD outperforms all baseline methods at all measured metrics except accuracy, where FCRO performs the best.} In terms of performance, FairREAD reaches the highest AUC in classifying all the diseases examined. In terms of fairness, FairREAD is consistently among the top 2 in all the examined methods. Notably, FairREAD outperforms other methods significantly at $FATE_{EO}$ and $FATE_{AUC}$, which demonstrates that our method achieves the best comprehensive result considering both performance and fairness.

\textbf{FairREAD demonstrates stability in performance across all examined diseases.} It consistently ranks in the top two methods for $FATE_{EO}$ and $FATE_{AUC}$ across all diseases. Furthermore, it is the only method with positive $FATE_{EO}$ and $FATE_{AUC}$ values for every \rev{pathology}, highlighting its robustness in enhancing the trade-off between performance and fairness.

\subsection{Ablation studies}\label{subsec:ablation}

\begin{table}[tbh]
\centering
\footnotesize
\caption{Performance of FairREAD averaged over all examined diseases with different threshold selection methods.}
\label{tab:threshold-methods}
\begin{tabular}{l c c c}
\toprule
\textbf{Method} & Accuracy\rev{($\uparrow$)} & $\Delta_{EO}$\rev{($\downarrow$)} & \rev{$FATE_{EO}$($\uparrow$)} \\
\midrule
\textbf{Min-gap} &
0.736 $\pm$ 0.010 & \textbf{0.129 $\pm$ 0.009} & \textbf{0.197} \\
\midrule
\textbf{Youden's J} & 0.736 $\pm$ 0.013 & 0.176 $\pm$ 0.019 & 0.055 \\
\midrule
\textbf{G-Means}     & 0.737 $\pm$ 0.011 & 0.155 $\pm$ 0.019 & 0.108 \\
\midrule
\textbf{Default}        & \textbf{0.871 $\pm$ 0.001} & 0.377 $\pm$ 0.035 & -0.493 \\
\bottomrule
\end{tabular}
\end{table}

\textbf{Ablation on threshold selection method.} We fit the classification threshold for FairREAD using different threshold selection methods, including Min-gap (ours), Youden's J, G-Means, and default threshold (where we use a classification threshold of 0.5). The results averaged over all four examined diseases are shown in Table \ref{tab:threshold-methods}. Note that threshold selection is independent to the parameters of the model, therefore here we fix the trained model and only re-fit the threshold using the training set, in order to prevent the randomness in the training process to affect comparison of threshold selection methods. Furthermore, since the threshold selection process does not affect threshold-agnostic metrics including $AUC$, $\Delta_{AUC}$, and $FATE_{AUC}$, these metrics are not shown in Table \ref{tab:threshold-methods}.

We can observe that Min-gap achieves the best performance in terms of $\Delta_{EO}$ and $FATE_{EO}$. Also, the accuracy of Min-gap is comparable to Youden's J and G-Means. The high accuracy of the default threshold is achieved trivially by taking advantage of the class imbalance in the dataset. Specifically, a default threshold value of 0.5 is often too high, leading FairREAD to classify almost all the images as negative. Since positive images only takes up a small proportion (as shown in Table \ref{tab:positive_rates}), the accuracy is superficially high.

\textbf{Ablation on adversarial learning.} In the training process of FairREAD, $\alpha_{adv}$ is a crucial hyperparameter that controls the weight of the adversarial loss in the second stage of training. Figure \ref{fig:ablation-alpha} shows the performance of FairREAD on Fracture classification with different $\alpha_{adv}$ values ranging from \rev{0} to 3. \rev{We can observe that when $\alpha_{adv}=0$, which means when adversarial learning is disabled, though the model has a higher accuracy, other metrics get significantly worse, especially $FATE_{EO}$ and $FATE_{AUC}$, which indicates that the model's performance-fairness tradeoff deteriorates.} 

\rev{Additionally,} we can observe that in general, the performance metrics (Accuracy, AUC) decrease as $\alpha_{adv}$ increases. This is because a larger $\alpha_{adv}$ value adds a stronger constraint on the model's optimization. Moreover, as $\alpha_{adv}$ increases, the comprehensive metrics ($FATE_{EO}$, $FATE_{AUC}$) first increases to a maximum point, and then decreases. This is because when $\alpha_{adv}$ is too small, FairREAD does not perform well on the fairness metrics ($\Delta_{AUC}$, $\Delta_{EO}$). On the other hand, when $\alpha_{adv}$ is too large, the model suffers from worse performance.

\subsection{Embedding space visualization}
To illuminate on how FairREAD balances fairness and classification performance, we employ t-distributed Stochastic Neighbor Embedding (t-SNE) \citep{t-sne} to visualize the embedding spaces at various stages of the model. Figure~\ref{fig:emb-combined} presents these visualizations for the Cardiomegaly classification task, highlighting the progression from the fair image encoder to the final fused representations. For comparison, we also include the embedding space of a standard Empirical Risk Minimization (ERM) model.

In the embedding space of $\zt$, the output of the fair image encoder, we can observe that samples from different demographic attributes do not form distinct clusters, indicating that $\zt$ is largely invariant to these sensitive attributes. This demonstrates the effectiveness of the disentanglement process, where the encoder successfully removes demographic information from the representation. However, when colored by target labels in Figure~\ref{fig:emb-target}, the positive and negative samples also become less clearly separable compared to the ERM model. This suggests a trade-off where enhancing fairness slightly compromises the immediate separability of target classes within $\zt$.

During the re-fusion process, the mean ($\mu$) and std ($\sigma^2$) that encodes the demographic attributes are first calculated. We can observe that as expected, the embedding space of mean and std is clearly separable in terms of demographic attributes. As the representations pass through the first and second re-fusion blocks, the demographic subgroups begin to re-emerge in the embedding space. Concurrently, the separation between target classes becomes more distinct compared to $\zt$. This indicates that the re-fusion mechanism successfully integrates demographic information and enhances classification performance.

\rev{\subsection{Out-of-distribution testing}}
\rev{Previous work \citep{ShortcutEncode} pointed our that conducting OOD tests is crucial for examining the effectiveness of bias mitigation methods in AI for medical imaging. 	To evaluate FairREAD’s performance under out-of-distribution (OOD) settings, we test models trained for Cardiomegaly and Pleural Effusion on the CheXpert dataset using the MIMIC-CXR dataset \citep{mimic-cxr} as an OOD test set for the same diseases. Note that Pneumonia and Fracture were not included in the test since there are too few samples with these diseases in the MIMIC-CXR test set for some subgroups that the calculation of accuracy and AUC is not reliable. The results are shown in Table~\ref{tab:ood_comparison}. We can observe that FairREAD is able to outperform all other baselines in terms of both performance and fairness, which proves its capability of being applied under this OOD setting.}

\begin{table*}[tbh!]
\centering
\footnotesize
\caption{\rev{Results of OOD testing on Cardiomegaly and Pleural Effusion. Best performance is marked in bold, second best performance is underlined.}}
\label{tab:ood_comparison}
{
\begin{tabular}{l l c c c c c c}
\toprule
\textbf{Pathology} & \textbf{Method} & Accuracy($\uparrow$) & AUC($\uparrow$) & \textbf{$\Delta_{AUC}$($\downarrow$)} & \textbf{$\Delta_{EO}$($\downarrow$)} & $FATE_{EO}$($\uparrow$) & $FATE_{AUC}$($\uparrow$) \\
\midrule
Cardiomegaly & ERM                    & 0.656 & 0.787 & \underline{0.132} & 0.471 & 0.000 & \underline{0.000} \\
\cmidrule(lr){2-8}
& FCRO                   & \underline{0.716} & \textbf{0.798} & 0.156 & 0.395 & 0.253 & -0.168 \\
\cmidrule(lr){2-8}
& LNTL                   & 0.676 & 0.772 & 0.134 & \textbf{0.359} & 0.268 & -0.034 \\
\cmidrule(lr){2-8}
& FairAdaBN              & 0.712 & 0.659 & 0.134 & 0.536 & -0.053 & -0.178 \\
\cmidrule(lr){2-8}
& AL                     & \textbf{0.722} & 0.770 & 0.155 & 0.380 & \underline{0.294} & -0.076 \\
\cmidrule(lr){2-8}
& \textbf{FairREAD (ours)} & \textbf{0.722} & \underline{0.790} & \textbf{0.113} & \underline{0.378} & \textbf{0.298} & \textbf{0.148} \\

\midrule
Pleural Effusion & ERM                & 0.773 & 0.904 & 0.140 & 0.304 & 0.000 & 0.000 \\
\cmidrule(lr){2-8}
& FCRO                   & \underline{0.813} & 0.905 & 0.127 & \underline{0.193} & \underline{0.417} & 0.094 \\
\cmidrule(lr){2-8}
& LNTL                   & 0.806 & 0.899 & \underline{0.115} & 0.254 & 0.207 & 0.173 \\
\cmidrule(lr){2-8}
& FairAdaBN              & 0.803 & 0.846 & \textbf{0.085} & 0.268 & 0.157 & \underline{0.329} \\
\cmidrule(lr){2-8}
& AL                     & 0.806 & \underline{0.910} & 0.142 & 0.229 & 0.289 & -0.008 \\
\cmidrule(lr){2-8}
& \textbf{FairREAD (ours)} & \textbf{0.823} & \textbf{0.911} & \textbf{0.085} & \textbf{0.166} & \textbf{0.519} & \textbf{0.401} \\

\bottomrule
\end{tabular}
}
\end{table*}

\section{Discussion}
In this paper, we propose a novel re-fusion after disentanglement (FairREAD) framework to mitigate unfairness in medical image classification without sacrificing performance of the model. To achieve this, we first train a fair image encoder that encodes the input image while discarding information with respect to demographic attributes. \rev{Concretely, we propose to combine orthogonality loss and adversarial training to ensure that all demographic information is removed from the image representation across the two stages of training. By doing so, we remove the potential shortcuts related to demographic attributes that might be utilized for diagnosis.} Next, we pass the fair image representation into a re-fusion module, where the encoded demographic attributes are fused with the disentangled image representation. \rev{This ensures that no potentially useful demographic information is lost.} Finally, we apply subgroup-specific decision threshold to the output logit to derive the classification output. Specifically, we propose the Min-gap strategy to decide the decision threshold for each subgroup.

\subsection{Experimental summaries}

Our experiments show that FairREAD consistently outperform the baseline methods across key metrics (Table~\ref{tab:results-main}). First of all, FairREAD achieves superior performance in terms of AUC, and performance comparable to other methods in terms of accuracy, showcasing its efficacy in maintaining high diagnostic performance. Secondly, FairREAD demonstrates significant reductions in fairness metrics, specifically disparity in Equal Odds ($\Delta_{EO}$) and AUC ($\Delta_{AUC}$), emphasizing its ability to mitigate bias effectively. Most importantly, FairREAD \rev{surpasses} all baseline methods with significant margins in terms of average $FATE_{EO}$ and $FATE_{AUC}$, which proves that its comprehensive performance in terms of both performance and fairness is the best among all compared methods. We also conduct ablation studies on two key components in FairREAD - subgroup-specific threshold (Table~\ref{tab:threshold-methods}) and adversarial learning (Figure~\ref{fig:ablation-alpha}). Results show that the Min-gap strategy we adopt in FairREAD achieves best performance in terms of $FATE_{EO}$ among potential threshold selection methods. Also, the weight of the adversarial loss ($\alpha_{adv}$) influence the trade-offs between performance and fairness metrics. Optimal settings of ($\alpha_{adv}$) are identified that best balance these aspects, measured by $FATE_{EO}$ and $FATE_{AUC}$. \rev{Additionally, we conducted OOD testing experiments to examine the performance of FairREAD under OOD settings, where the model trained on CheXpert is tested on MIMIC-CXR. Our results indicate that FairREAD is able to outperform baseline methods in this setting, which further demonstrates the robustness of the FairREAD model.}

\subsection{Limitations and future work}
The presented work has opportunities for improvement, which are the subject of our ongoing work. In the following, we summarize some potential limitations of our work and potential opportunities for future development and extension.

\textbf{\rev{Binarization} of demographic attributes.} In this paper, we discretize each of the three demographic attributes (age, gender, race) into binary values for simplification. This is mainly because a more detailed separation of demographic subgroups can reduce the number of samples within each subgroup, and further \rev{hinders} the \rev{accurate} calculation of metrics. \rev{For example, there are only 31 Native American samples within our curated CheXpert dataset, and the test set only contains 4, which makes it difficult to reliably calculate accuracy and AUROC, and therefore also $\Delta_{EO}$ and $\Delta_{AUC}$. However,} though it is a common practice in many previous works involving group fairness metrics (e.g. \cite{FairOrth, FairAdaBN}), binarization of demographic attributes can conceal potential unfairness within each subgroup\rev{, and also undermines the value of fairness-related research by reducing diversity in the dataset}. We leave for future work to examine \rev{whether} the \rev{FairREAD architecture} can generalize to more fine-grained separations of subgroups\rev{, particularly when larger datasets for currently underrepresented demographics become available}.

\textbf{Artificial splitting of demographic subgroups} Following previous works in unfairness mitigation for medical imaging \citep{Yang2024, Chexclusion, cardiac-mri-seg-fairness}, we apply $n$ demographic attributes to split all training samples into $2^n$ groups ($n=3$ in our case). However, such a method to split training samples can hardly be exhaustive, since there are too many demographic attributes to potentially consider (e.g. maritial status, gender identity, sexual orientation, etc.) A potential direction for improvement is to add individual fairness metrics \citep{dwork2011fairnessawareness} into consideration, which we leave for future work to explore.

\textbf{Combining FairREAD with resampling.} In this paper, we use threshold-dependent metrics such as Accuracy, $\Delta_{EO}$, and $FATE_{EO}$ to measure the performance of FairREAD. In order to make sure the classification threshold is selected appropriately, we employ the Min-gap strategy to determine a specific threshold value for each demographic subgroup. However, there are other potential strategies to resolve this issue, such as dataset resampling \citep{resample}. It would be a interesting topic to explore how dataset resampling influence the performance and fairness of FairREAD, and whether combining dataset resampling and subgroup-specific threshold could improve the model further.

\textbf{Applying the re-fusion after disentanglement mechanism with other fair representation learning methods.} The general architecture of FairREAD is generalizable to other methods to construct the fair image encoder. \rev{For example, many methods based on adversarial learning \citep{adv, LNTL, li2021estimatingimprovingfairnessadversarial}, disentanglement learning \citep{Orth}, contrastive learning \citep{fairdisco}, pruning \citep{wu2022fairpruneachievingfairnesspruning, jung2024unifieddebiasingapproachvisionlanguage}, etc., have been proven effective in removing information related to demographic attributes from the image representation. Future technical improvements in fair image encoder construction could potentially further improve FairREAD's performance.}

\textbf{Applying other modality fusion methods.} In the re-fusion mechanism introduced in Section \ref{subsec:refusion}, we achieved modality fusion between demographic attributes and image using a rescaling-based technique, where the demographic attributes are encoded into mean and variance applied upon the projected image representation vector. Other modality fusion methods, such as cross-attention have the potential to replace the current re-fusion strategy, which we important direction for our future explorations.

\section{Conclusion}

This paper presents FairREAD, a novel framework designed to address the fairness challenges in medical image classification without compromising diagnostic performance. Through the integration of disentanglement learning with a novel re-fusion mechanism, FairREAD achieves a balance between fairness and performance by leveraging sensitive demographic attributes reasonably. The subgroup-specific threshold adjustment further enhances the fairness of predictions, demonstrating a significant reduction in performance disparities across demographic subgroups. The extensive evaluation of FairREAD on the CheXpert dataset highlights its ability to achieve significant improvement of fairness-performance trade-off robustly across various pathologies. 

\bibliographystyle{model2-names.bst}\biboptions{authoryear}
\bibliography{refs}
\clearpage
\appendix
\rev{\section{Details about Hyperparameter Searching}\label{app:hyperparamter}}

\rev{The hyperparameters in FairREAD include: adversarial learning weight $\alpha_{adv}$, number of re-fusion blocks ($N$), hidden dimension in each re-fusion block ($d_{hidden}$), dropout rate, as well as orthogonal loss weights $\lambda_c$ and $\lambda_r$. For $\lambda_c$ and $\lambda_r$, we directly use the values suggested by \citet{FairOrth}. For other parameters, we use a grid search strategy to find the optimal value. Specifically, we conduct five-fold cross-validation as described in Section~\ref{subsec:dataset} for each set of parameters. We then choose the best set of parameters in terms of average validation set performance. All tested values are shown in Table~\ref{tab:hyperparam_grid}, and the selected hyperparameters are shown in Table~\ref{tab:hyperparam_selected}.}
\begin{table*}[!t]
  \centering
  \caption{\rev{Grid‐search hyperparameter values.}}
  \label{tab:hyperparam_grid}
  \begin{tabular}{ll}
    \toprule
    Hyperparameter & Tested values \\
    \midrule
    Adversarial weight ($\alpha_{adv}$)        & $\{0,\,0.1,\,0.3,\,0.5,\,0.8,\,1,\,2\}$ \\
    \# of re‐fusion blocks ($N$)                  & $\{1,\,2,\,3\}$ \\
    Hidden dimension in each re‐fusion block ($d_{hidden}$) & $\{256,\,1024\}$ \\
    Dropout rate                             & $\{0.1,\,0.3\}$ \\
    \bottomrule
  \end{tabular}
\end{table*}

\begin{table*}[!htb]
  \centering
  \caption{\rev{Selected hyperparameters for final models on MIMIC.}}
  \label{tab:hyperparam_selected}
  \begin{tabular}{lcccc}
    \toprule
    Target            & $\alpha_{adv}$ & $N$ & $d_{hidden}$ & Dropout rate\\
    \midrule
    Cardiomegaly      & 0.1            & 2         & 256        & 0.1     \\
    Pleural Effusion  & 0.1            & 2         & 256        & 0.1     \\
    Pneumonia         & 0.5            & 2         & 256        & 0.1     \\
    Fracture          & 0.5            & 1         & 256        & 0.1     \\
    \bottomrule
  \end{tabular}
\end{table*}

\end{document}